\documentclass{article}

\usepackage{main}

\usepackage[utf8]{inputenc} 
\usepackage[T1]{fontenc}    
\usepackage{hyperref}      
\usepackage{url}            
\usepackage{booktabs}       
\usepackage{amsfonts}       
\usepackage{nicefrac}       
\usepackage{microtype}      
\usepackage{xcolor}         
\usepackage{graphicx}
\usepackage{multirow}
\usepackage{amsmath}
\usepackage{multicol}
\usepackage{float}

\title{{SkyReels-A2}: Compose Anything in Video Diffusion Transformers}

\author{%
Zhengcong Fei, \  Debang Li,  \ Di Qiu
\\
Kunlun Inc.\\
Beijing, China\\
{\tt\small \{author\}@gmail.com}
}

\begin{document}

\maketitle
\vspace{-0.5cm}
\begin{figure}[H]
  \includegraphics[width=\textwidth]{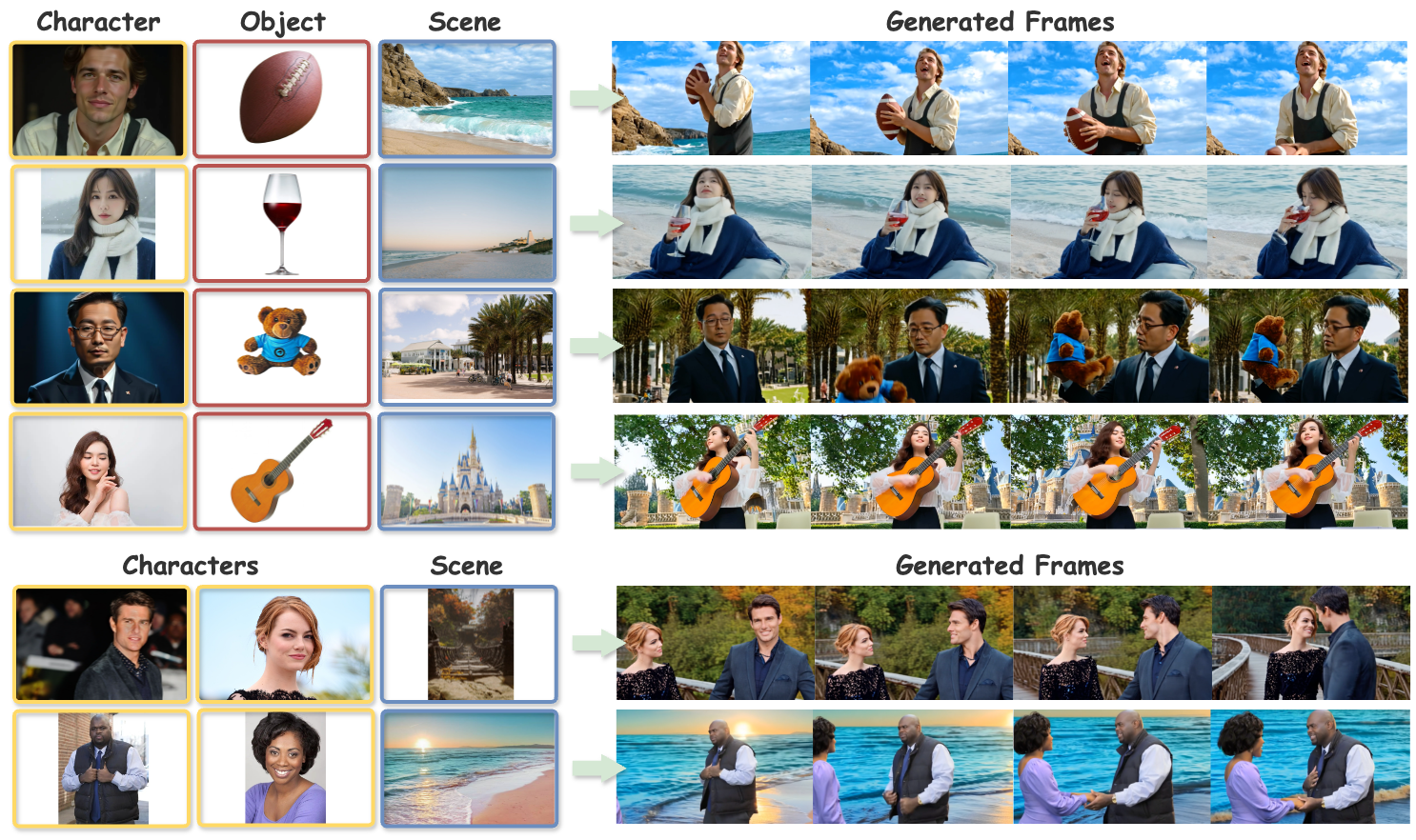}
  \caption{\textbf{Examples of \emph{elements-to-video} results from our proposed \texttt{SkyReels-A2} model.} Given reference with multiple images and textual prompt, our method can generate realistic and naturally composed videos while preserving specific identity consistent. 
  }
  \label{fig:case1} 
\end{figure}

\renewcommand{\thefootnote}{} 
\footnotetext{~$^*$Equal Contributions}
\footnotetext{~$^\dagger$Project Lead}
\begin{abstract}
  This paper presents \texttt{SkyReels-A2}, a controllable video generation framework capable of assembling arbitrary visual elements (e.g., characters, objects, backgrounds) into synthesized videos based on textual prompts while maintaining strict consistency with reference images for each element. 
  We term this task \emph{elements-to-video (E2V)}, whose primary challenges lie in preserving the fidelity of each reference element, ensuring coherent composition of the scene, and achieving natural outputs. 
  To address these, we first design a comprehensive data pipeline to construct prompt-reference-video triplets for model training. 
  Next, we propose a novel image-text joint embedding model to inject multi-element representations into the generative process, balancing element-specific consistency with global coherence and text alignment. 
  We also optimize the inference pipeline for both speed and output stability. 
  Moreover, we introduce a carefully curated benchmark for systematic evaluation, i.e, \texttt{A2 Bench}. 
  Experiments demonstrate that our framework can generate diverse, high-quality videos with precise element control.
  \texttt{SkyReels-A2} is the first open-source commercial grade model for the generation of \emph{E2V}, performing favorably against advanced closed-source commercial models.
  We anticipate \texttt{SkyReels-A2} will advance creative applications such as drama and virtual e-commerce, pushing the boundaries of controllable video generation.
  The code and model weights of \texttt{SkyReels-A2} are publicly available at \url{https://github.com/SkyworkAI/SkyReels-A2}.
\end{abstract}

\section{Introduction}

The emergence of diffusion models has significantly advanced the field of generative modeling, fundamentally transforming the paradigm of content creation~\cite{DDIM,rombach2022stablediffusion,peebles2023dit,esser2024scaling,fei2024scaling,fei2024dimba}.
Building upon text-to-image architectures, large-scale pre-trained video diffusion frameworks have achieved remarkable breakthroughs by incorporating temporal consistency mechanisms into their attention designs~\cite{hong2022cogvideo,blattmann2023stable,singer2022makeavideo,girdhar2023emuvideo,svd}. 
Notably, the scalability of diffusion transformer architectures with full 3D attention~\cite{peebles2023dit,opensora_plan, opensora, allegro, yang2024cogvideox,movie_gen,kong2024hunyuanvideo,wan2025} has propelled the field toward high-quality visual content, enabling a broad spectrum of downstream applications~\cite{magictime, controlnet, InstaDrag, evagaussians, cycle3d, ViewCrafter,fei2024video}. 
These controllable scenes are impactful in the area of limited subject customization~\cite{dreamvideo, customvideo, motionbooth, Still-moving, magic-me, fang2024motioncharacter}, facilitating highly customizable and dynamic visual outputs~\cite{fei2025ingredients,qiu2025skyreels}. 
Despite these remarkable advancements, releasing video diffusion model capacity into multiple consistent elements generation~\cite{xing2024survey}, such as AI drama and virtual e-commerce, remains an ongoing challenge.


Contemporary foundational video generation models predominantly address two core tasks: text-to-video (T2V) and image-to-video (I2V). 
T2V usually utilizes T5~\cite{raffel2019exploring} or CLIP~\cite{radford2021learning} text encoder to interpret textual instructions and generate visual content that reflects the desired characters, actions, and settings. 
Although T2V fosters creative and diverse content generation, it frequently struggles to maintain consistent and predictable outcomes due to the inherent stochasticity of diffusion process~\cite{movie_gen,yang2024cogvideox}. 
Conversely, I2V typically transforms a static image into a dynamic video by providing an initial frame alongside optional textual descriptions, which is often constrained by the "copy-paste" dependency on the initial frame~\cite{yuan2024identity,fei2025ingredients,huang2025conceptmaster,liu2025phantom}.
In this paper, we endeavor to visual \emph{elements-to-video} task, that synthesize dynamic and natural composed videos that precisely follows multiple reference images and textual prompts, thereby integrating the flexibility of text-based generation with the controllability in image conditioning. 
It is worth noting that the scope of referenced compositions are not limited to human characters or subjects, but also encompass any animals, landscapes, and various other visual elements.

In this work, we first construct a data structure consisting of text-reference-video triplets, where reference includes multiple images for different visual elements.
We have re-annotated and aligned the video captions to focus on describing the appearance and action of the elements in the video. 
Additionally, the reference images for elements are not naively taken from a single video frame but are sampled and selected across multiple videos, ensuring that the generated video does not simply copy and paste the images. 
Then, we develop a unified video generation framework for any visual element, which is competitive with existing solutions on the market. 
Specifically, we design a joint image-text injection from semantic and spatial ways based on existing foundational video models to ensure effective learning of cross-modal data representations. 
Furthermore, we propose \texttt{A2-Bench}, a novel benchmark for comprehensively evaluating the \emph{E2V} task, which exhibits statistically significant correlation with human subjective judgments. 
Finally, we employ a training-free acceleration for user-friendly experience.
Extensive experiments demonstrate the superiority of our proposed method, highlighting its efficacy as a generative control mechanism. 

In summary, our methodology demonstrates robust flexibility, enabling the seamless integration of diverse visual elements within the synthesized video outputs, making it highly adaptable to diverse applications. 
The contributions are summarized as below: 
\begin{itemize}
    \item We propose \texttt{SkyReels-A2}, an \emph{elements-to-video (E2V)} framework based on video diffusion models, aiming to maintain fidelity from multiple reference images (characters, objects, and background) while enabling precise control through textual instructions.
    \item We introduce a meticulously curated dataset comprising high-quality text-reference-video triplets, and develop \texttt{A2-Bench} which facilitates comprehensive and automated assessment of \emph{elements-to-video} tasks.
    \item Extensive experiments demonstrate that \texttt{SkyReels-A2} can generate high-quality, editable, and temporally consistent multi-visual-elements videos, performing favorably against advanced commercial closed-source models in both qualitative and quantitative analyses. To encourage further advancements in this domain, we have publicly released our code, models, and evaluation benchmark.
\end{itemize}

\section{Methodology}

Formally, given a set of $N$ input reference images, denoted as $\{\mathcal{C}_n\}_{n=1}^N$, describe $N-1$ individual subjects and background of an image, our goal is to generate high-quality and natural videos composed of these visual elements while following text prompts.
We first discuss text-to-video diffusion models and image condition preliminaries in Section 2.1. Following this, we details architecture of \texttt{SkyReels-A2} for composition video in Section 2.2. Finally, we describe dataset construction pipeline in Section 2.3 and evaluation \texttt{A2-Bench} in Section 2.4, respectively.

\subsection{Preliminaries}

\paragraph{Text-to-video diffusion.}

Diffusion-based text-to-video models usually learns to iteratively transforms a noise $\epsilon$ into a video $x_0$. 
Early approaches implement the denoising process directly within the pixel space \cite{DDIM, DDPM, DPM}. However, due to the substantial computational costs, more recent techniques predominantly leverage latent spaces \cite{LDM, kondratyuk2023videopoet, magictime, yang2024cogvideox,kong2024hunyuanvideo} with 3D VAE. 
The optimization objective can be defined as:
\begin{equation}
\label{eq: original_loss_function}
{L}_{diff}=\mathbb{E}_{x_0, t, y, \epsilon}\left[\left\|\epsilon-\epsilon_\theta\left(\mathcal{E}\left(x_0\right), t, \tau_\theta(y)\right)\right\|_2^2\right]
\end{equation}
where $y$ denotes textual prompt, $\epsilon$ is randomly sampled from a Gaussian distribution, \emph{e.g.}, $\epsilon \sim \mathcal{N}(0, 1)$, and $\tau_\theta(\cdot)$ is the text encoder, e.g., T5 \cite{raffel2019exploring}. Substituting $x_0$ with its latent representation $\mathcal{E}\left(x_0\right)$, form the diffusion process. Note that our method following more advanced flowing matching \cite{esser2024scaling} to optimize the velocity with MSE loss. 
Video generation models based on DiT demonstrates considerable potential in capturing the dynamics of the physical world \cite{SORA, yang2024cogvideox, allegro}. Despite the recent advancements achieved through scaling along 1D sequences, research on controllable generation within this framework remains relatively under-explored. 
In particular, no prior study has investigated the mechanisms underlying the efficacy of DiT in scenarios of multi-visual-elements composition. 
Conversely, commercial models have consistently demonstrated superior performance in this domain. This work aims to bridge the gap between open-source methodologies and proprietary solutions, addressing this unexplored aspect to advance the state of controllable video generation.

\paragraph{Condition with images.}

Although natural language provides an intuitive means for controlling generative models, it often lacks the specificity required to accurately describe objects. To address this limitation, recent approaches \cite{kong2024hunyuanvideo,hong2022cogvideo} extend text-conditioned diffusion models to condition on reference images. For example, in Wan \cite{wan2025}, the image prompt $\mathcal{I}_{img}$ is first encoded with a pre-trained image encoder, i.e., CLIP, to obtain visual embeddings $E_{vision}(\mathcal{I}_{img})$, and then the hidden states $\mathcal{H}_{img}$ are projected to the key matrix $f_K^{img}(\mathcal{H}_{img})$ and values matrix $f_V^{img}(\mathcal{H}_{img})$. Analogous to text prompts, the image prompt is integrate with additional cross-attention for queries of video sequence as:
\begin{equation}
 \text{Softmax}(\frac{f_Q^{text}(H_{txt}) \cdot f_K^{img}(\mathcal{H}_{img})^T}{\sqrt{d}}) \cdot f_V^{img}(\mathcal{H}_{img}).
\end{equation}
The final output hidden representation is obtained by summing the outputs of text-prompt cross-attention and image-prompt-based attention.
Meantime, reference image $\mathcal{I}_{img}$ is also encoded with VAE to obtain spacial latent features, which is concatenate with input noise in channel dimension before forward into patch embedding. 
Different image-conditioned methods \cite{hong2022cogvideo,kong2024hunyuanvideo} differ in encoder design and image integration branches.

\subsection{Architecture}
\begin{figure*}[t]
  \centering
   \includegraphics[width=1\linewidth]{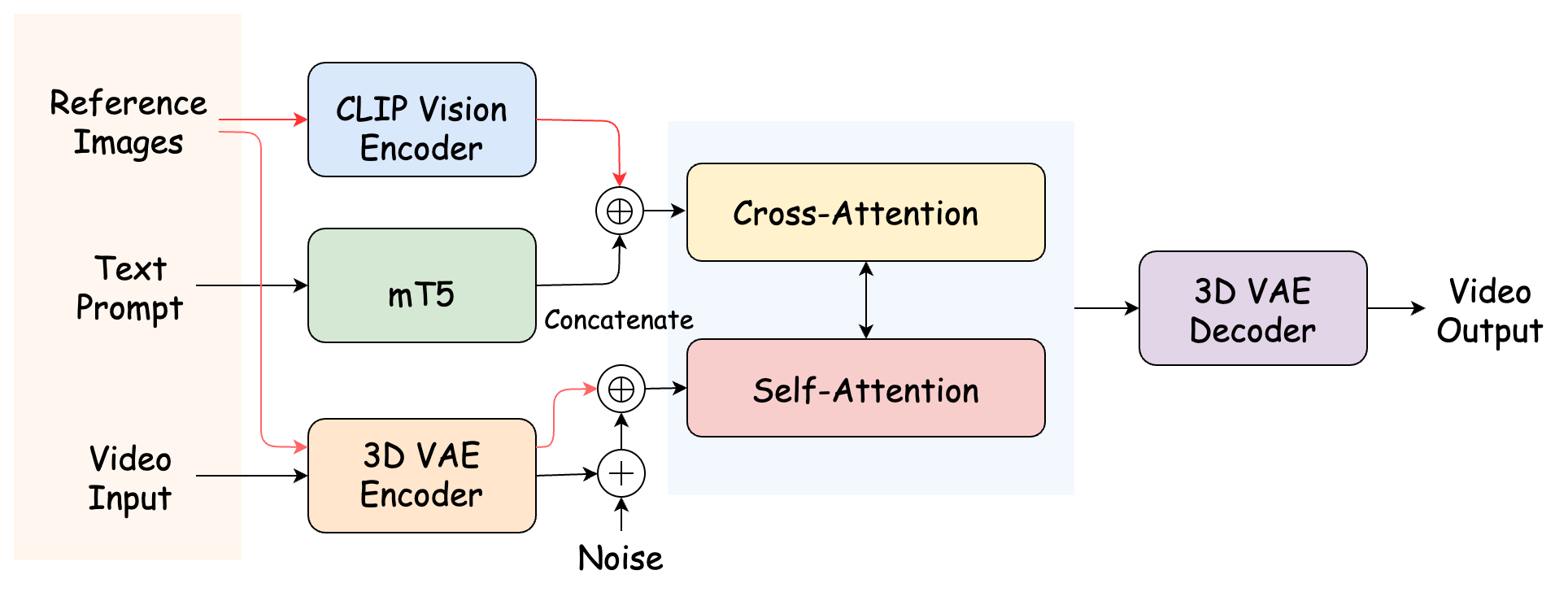}
   \caption{\textbf{Overview of \texttt{Skyreels-A2} framework.} Our approach initiates by encoding all reference images using two distinct branches. The first, termed the spatial feature branch (represented in red bottom arrow), leverages a fine-grained 3D VAE encoder to process per-composition images. The second, identified as the semantic feature branch (represented in red top arrow), utilizes a CLIP vision encoder followed by an MLP projection to encode semantic references. Subsequently, the spatial features are concatenated with the noised video tokens along the channel dimension before being passed through the diffusion transformer blocks. Meanwhile, the semantic features extracted from the reference images are incorporated into the diffusion transformers via supplementary cross-attention layers, ensuring that the semantic context is effectively integrated during diffusion.
   }
   \label{fig:framework} 
\end{figure*}

The overall architecture is depicted in Figure~\ref{fig:framework}.
Given a set of $N$ reference composed images and background images, denoted as $\{\mathcal{C}_n\}_{n=1}^N$, we aim to generate high-quality videos that preserve the identity of each reference and background, while following textual prompt $\mathcal{T}$ with flexible layouts.
Our approach leverages an advanced video diffusion transformer architecture, with minimal structural modifications to ensure adaptability across a broad spectrum of applications.
We segment each subjects in the reference image with white grounding except background reference image, to avoid additional noise. 
Each input reference subject, $\mathcal{C}_n$, is processed through a two-stream structure for cross-modal projection. The first stream use a semantic image encoder $E_{img}$ to extract visual embedding for global and semantic features.  The second stream use original VAE to obtain spacial and local detailed features.

Specifically, the semantic encoder is instantiated using a CLIP image encoder, which extracts grid-based features from its penultimate layer. A projection module subsequently transforms these features into image queries with dimensions aligned to the video sequence queries. The resulting image tokens from all reference images are concatenated and utilized as keys and values within the cross-attention layers, which are integrated after each text-prompt cross-attention block.
For the spatial branch, reference images are first concatenated along the frame dimension and zero-padded to match the original frame number. A standard 3D VAE is then applied to extract video latents, which are subsequently concatenated with the noise latents along the channel dimension before being passed through the patch embedding module.

\subsection{Training and Inference}

\noindent \textbf{Training objectives.}
Given the composition prompts $\{\mathcal{C}_n\}_{n=1}^N$ and input text prompt $\mathcal{T}$, we first segment the conception in original images without background, train our \texttt{SkyReels-A2} model to reconstruct the target video in the latent space. Our training objective follows standard diffusion MSE losses as Equation~\ref{eq: original_loss_function}. During training, we only optimize the following neural modules: cross-attention, patch embedding, image condtion embedders, while keeping the remain parts frozen. 

\noindent \textbf{Inference acceleration.} 
Following \cite{wan2025}, we adopt UniPC multi-step schedule \cite{zhao2023unipc} for inference sampling. Meantime, we also consider some effective and advanced acceleration strategy\footnote{https://github.com/chengzeyi/ParaAttention} for better practical employment. 
Generally, parallelizing the processing of neural network activations across multiple GPUs during diffusion model inference is a crucial acceleration strategy, particularly as our model scales to 14B parameters, where slow inference in each sampling step becomes a bottleneck. Our method employs Context Parallel, CFG Parallel, and VAE Parallel strategies to enhance model efficiency, enabling rapid and lossless video generation while meeting the stringent low-latency requirements of online environments. 
Additionally, we implement User-Level GPU Deployment, leveraging model quantization and parameter-level offload strategies to significantly reduce GPU memory consumption, thereby accommodating consumer-grade graphics cards with limited VRAM.

\subsection{Dataset Construction}
\begin{figure}[t]
  \centering
   \includegraphics[width=0.99\linewidth]{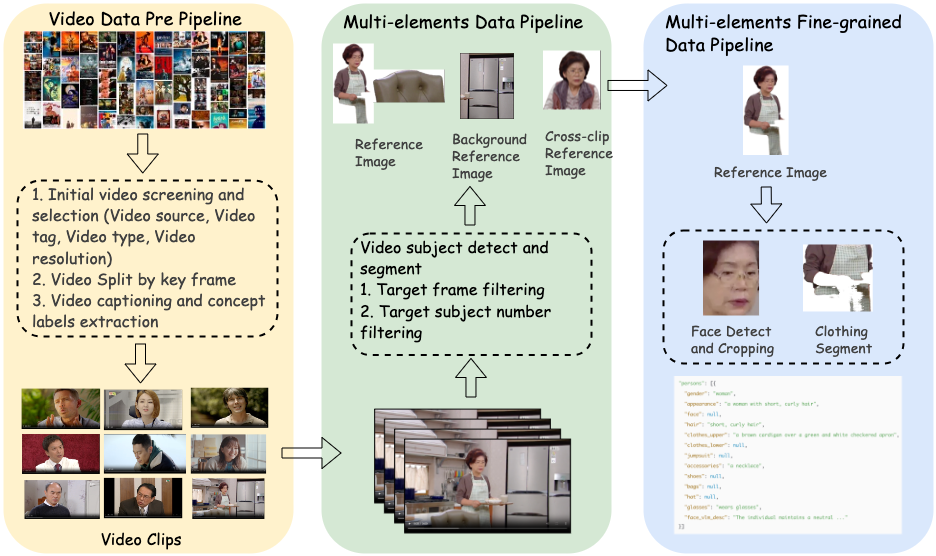}
   \caption{\textbf{Data processing pipeline in \texttt{SkyReels-A2}.} 
   The pipeline begins with preprocessing, where raw videos are filtered by resolution, labels, types, and sources, followed by temporal segmentation based on key-frames. 
   Next, a proprietary multi-expert video captioning model generates both holistic descriptions for video clips and structured concept annotations.
   Subsequently, detection and segmentation models extract visual elements (e.g., humans, objects, environments). 
   To mitigate duplication, reference images are retrieved from other clips based on clip/facial similarity score. 
   Further refinement includes face detection and human parsing to obtain facial/attire elements. 
   Finally, the extracted elements are matched with structured descriptions to form training triplets (visual elements, video clips, and textual captions).}
   \label{fig:data} 
\end{figure}
Data construction plays a pivotal role in achieving multi-subject consistent controllable video generation.
Unlike conventional text-to-video (T2V) or image-to-video (I2V) tasks, our framework requires additional reference images for various subjects (e.g., humans, objects, and scenes). 
To address this challenge, we design a comprehensive training data pipeline capable of generating high-quality video-caption-multi-reference triplets, which is illustrated in Figure~\ref{fig:data}.

Our pipeline begins with the collection of a large-scale video dataset. 
Each video is segmented into clips based on content coherence. 
We then employ an in-house multi-expert video captioning model to generate both holistic and structured captions for these clips. 
The structured captions encapsulate fine-grained details, including distinct subjects (e.g., humans, clothing, and objects), background information, as well as dynamic attributes such as facial expressions, actions, and motion trajectories.

Next, we proceed to construct reference images. For each video clip, a generic detection model~\cite{yolo11_ultralytics} is first applied to localize humans and objects. 
For human subjects, we further utilize a face detector~\cite{Deng2020CVPR} and a human parsing model~\cite{khirodkar2024sapiens} to extract facial features and apparel details. 
To align the detected subjects with the structured captions, we leverage the CLIP model~\cite{ilharco_gabriel_2021_5143773} to match textual descriptions in the captions with visual entities.

To mitigate the "copy-paste" effect in generated outputs, we introduce an additional similarity-based filtering step. 
Specifically, we compute inter-clip similarity for detected subjects using a face similarity model~\cite{Deng2020CVPR} (for humans) and a CLIP-based similarity model~\cite{ilharco_gabriel_2021_5143773} (for objects). 
This allows us to select diverse reference images for the same subject across different clips.
For background reference construction, we identify the frame with maximal background coverage, remove foreground objects via cropping, and retain the purified background image.

In conclusion, we compiled a dataset of approximately 2 million high-quality video-reference-prompt triplets for training purposes.

\subsection{A2-Bench Evaluation}
Existing video generation evaluation benchmarks, VBench\cite{huang2024vbench} and VBench++\cite{huang2024vbench++}, have introduced a rigorous evaluation framework for text-to-video and image-to-video tasks through a well-designed assessment suite and multi-dimensional evaluation criteria. However, a comprehensive benchmark remains lacking for downstream video generation task, elements-to-video. In order to thoroughly assess the performance of E2V task across diverse scenarios, we propose an automated, comprehensive, and human-aligned \texttt{A2-Bench}, which provides a systematic evaluation framework for assessing composition video generation models across multiple dimensions, ensuring rigorous and reliable performance measurement.

We collected 150 reference images from various scenarios as elements for our E2V task, comprising 50 distinct human identities, 50 different objects spanning 12 categories, and 50 unique backgrounds. To construct the benchmark dataset, we randomly paired multiple elements (character, object, and background) into 50 diverse input combinations. Corresponding text prompts were then generated using LLMs to facilitate the dataset creation process.It is important to note that we have meticulously ensured that there is no overlap between the training videos and the constructed \texttt{A2-Bench}.

The automatic metrics of \texttt{A2-Bench} comprise three core dimensions: composition consistency, visual quality, and prompt following, evaluated through eight fine-grained metrics.

\noindent \textbf{Composition Consistency.} Composition consistency is designed as the core metric for evaluating the consistency of video-generated elements in the E2V task. 
\begin{itemize}
    \item 
\emph{Character ID consistency} is used to evaluate the consistency of characters, with a straightforward evaluation approach. After detecting the face, a face recognition model\cite{deng2019arcface} is employed to extract features and measure cosine similarity. 
\item \emph{Object consistency} assesses the consistency of non-character objects, where we use Grounded-SAM to segment the object parts in the video and compute the similarity between frame-level clip features. 

\item \emph{Background consistency} measures the similarity between the generated video scene and the reference background image. This is done by detecting and segmenting the subject, masking out the subject, and then calculating the similarity of frame-level clip features with the reference background image.

\end{itemize}

\noindent \textbf{Visual Quality.}
To assess the visual quality of the generated videos, we incorporate comprehensive image quality, aesthetic quality, motion smoothness and dynamic degree, as defined by VBench\cite{huang2024vbench}. These metrics collectively capture both the temporal coherence and visual appeal of the generated content, ensuring a comprehensive evaluation of video quality.

\noindent \textbf{Prompt Following.}
We leverage ViCLIP to compute the cosine similarity score between textual descriptions and video content, providing a direct measure of text-video alignment. This approach enables an effective evaluation of the semantic consistency between the textual input and the corresponding visual representation.

\noindent \textbf{Comprehensive Score.}
We incorporate human feedback to derive a holistic assessment of composition consistency, visual quality and prompt following, recognizing that each dimension contributes differently to user preferences rather than applying a simple average.

\noindent \textbf{User Preference Study.} 
For the elements-to-video task, given the high error rates in automated element detection and matching, we conduct a user preference study to assess visual quality and element fidelity to complement automated evaluation. We use 50 test samples and present the conditional image, prompt, and results from multiple models (including Keling, Vidu, Pika, and our SkyReels-A2) to multiple participants. For each sample, participants simultaneously view the four results and rate them on a scale of 1 to 5 based on different evaluation criteria. Our approach of user study provides a more intuitive comparison of the performance differences between models in terms of visual quality and element details.

Notably, our user preference study adopts a highly detailed evaluation framework, encompassing 10 distinct criteria, e.g., instruction following, face consistency, spatial rationality, and subject coherence. Each sample is assessed through human ratings based on these fine-grained dimensions, ensuring a more precise evaluation of model performance. Additionally, we provide the full scoring guidelines in the Table \ref{tab:dimensions} to enhance transparency and facilitate a more rigorous and effective assessment of the element-to-video task.






\begin{figure*}[t]
  \centering
   \includegraphics[width=1\linewidth]{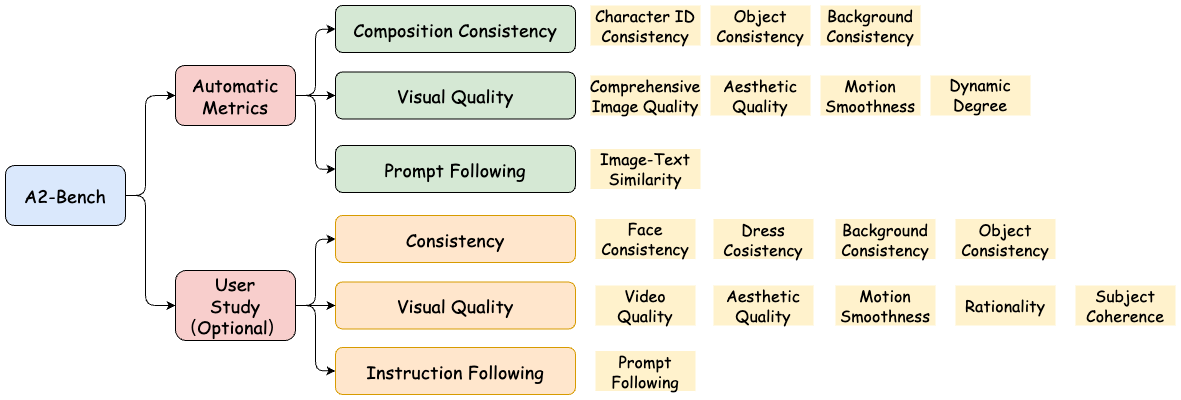}
   \caption{\textbf{The dimensions covered in \texttt{A2-Bench}. } Our evaluation consider both automatic metrics and user study, meantime, it covers multiple perspectives that precisely reflects the quality of \emph{E2V} task. } 
   \label{fig:bench} 
\end{figure*}

\section{Experiments}

\subsection{Experimental Settings}

\paragraph{Implementation details.}

\texttt{SkyReels-A2} is fine-tuned from a video generation foundation model based on the DiT architecture \cite{wan2025}. 
The T2V and I2V pre-training stages are excluded from this evaluation. We focus on assessing the elements-to-video generation capability, including character, object and background similarity.
During training, we drop the video captions and reference conditions with probabilities of 30\% and 10\% for classifier-free guidance, respectively. We pad with white image when the reference image is not equal to video ratios and the training video segments consist of 81 frames, corresponding to a duration of 6 seconds at 15 frames per second (FPS). The training process employs the Adam optimizer with a learning rate set to 1e-5 and a global batch size of 256. During inference, we utilize 50 steps and set the CFG scale to 5.

\paragraph{Baselines.}

For the elements-to-video task, the currently available state-of-the-art (SoTA) methods are closed source commercial tools. Therefore, we evaluate and compare the latest capabilities of Pika\cite{Pika}, Vidu\cite{Vidu}, and Keling\cite{Keling} products.



\begin{figure*}[t]
  \centering
   \includegraphics[width=1\linewidth]{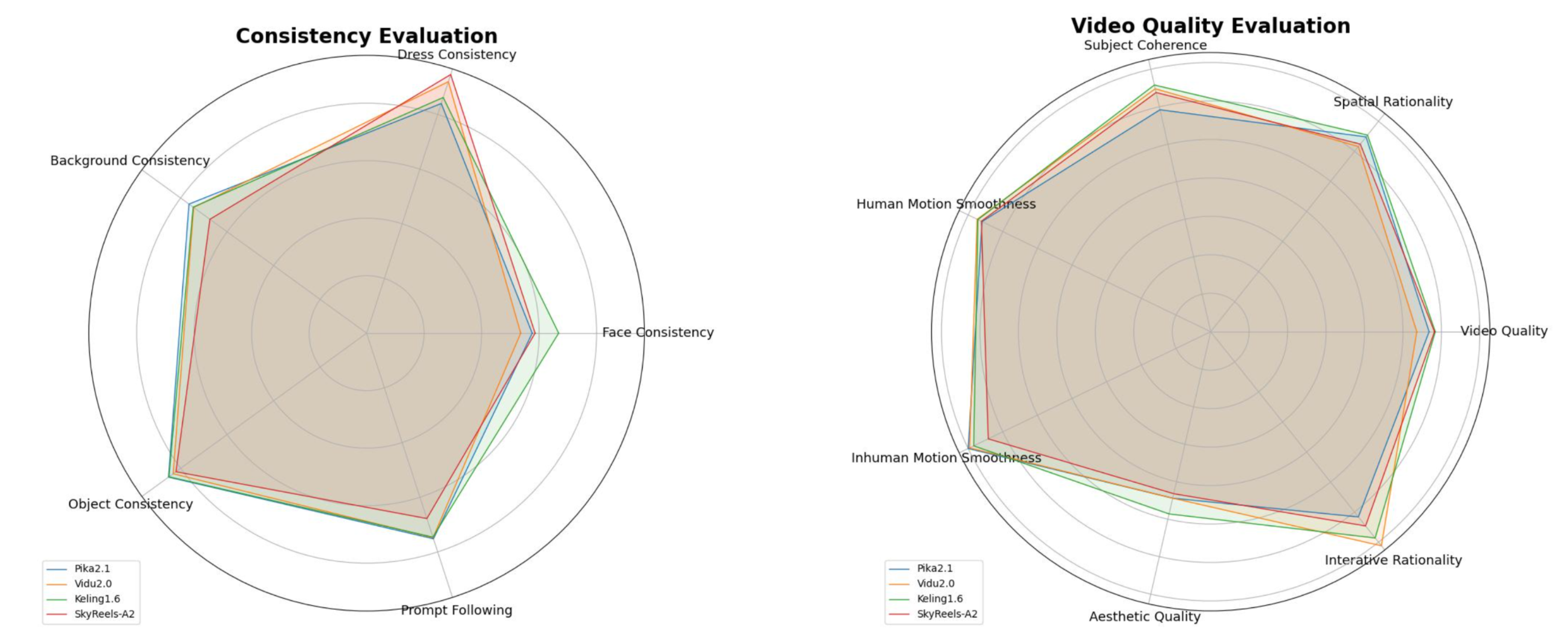}
   \caption{\textbf{Consistency and video quality of user study results for elements-to-video generation.} We can see that our \texttt{SkyReels-A2} achieve comparable generative performance compared with top-tier closed source commercial video models.}
   \label{fig:rada} 
\end{figure*}


\begin{table}[!t]
	\centering
	\setlength{\tabcolsep}{10pt}
    \caption{\textbf{Comparison of different methods based on the automatic metrics of \texttt{A2-Bench} Evaluation Dimension.} It contains three major dimensions including composition consistency, visual quality, and prompt following. We can see that our \texttt{SkyReels-A2} achieve comparable composition performance, especially in objective consistency. }
	\begin{tabular}{lcccc}
		\toprule
		\multicolumn{1}{l}{\textbf{Dimension}} & \multicolumn{1}{c}{\textbf{Pika2.1}} & \multicolumn{1}{c}{\textbf{Vidu2.1}} & \multicolumn{1}{c}{\textbf{Keling1.6}} & \multicolumn{1}{c}{\textbf{SkyReels-A2}}\\
		\midrule
		ID Consistency  & 0.388 & 0.455 & \textbf{0.497} & 0.398      \\ 
		Object Consistency   & 0.788 & 0.796 & 0.790 & \textbf{0.809}   \\ 
		Background Consistency   & \textbf{0.733} & 0.727 & 0.700 & 0.677 \\
		\midrule
		Image Quality  & 0.548 & 0.565 & 0.587 & \textbf{0.683}        \\
		Aesthetic Quality   & \textbf{0.628} & 0.616 & 0.609 &   0.579   \\
		Motion Smoothness   & 0.917 & \textbf{0.960} & 0.879 &   0.891    \\
		Dynamic Degree    & 0.918 & 0.878 & \textbf{1.000} &  \textbf{1.000}  \\
		Image-Text Similarity  & \textbf{29.128} & 28.926 & 27.504 & 28.188   \\
		\midrule
		\textbf{Comprehensive Score} & 0.807 & 0.821 & \textbf{0.826} & 0.818 \\
		\bottomrule
	\end{tabular}
	\label{tab:main}
\end{table}


\subsection{Quantitative Results}

The video quality evaluation results, are listed in the Table \ref{tab:main}. We can see that for visual consistency perspective, \texttt{SkyReels-A2} performs lightly poor on the background consistency metrics, while excelling in other metrics such object and character consistency. Moreover, in right part of video quality, \texttt{SkyReels-A2} leads in overall comparable metrics for dynamic degree and image quality. 
For the elements-to-video task, due to some error rates in automated subject detection and matching, we also conducted a user study. 
We surveyed multiple users, who rated the methods on a scale of 1 to 5 (The score larger, the results in human perception better). The evaluation results, displayed in the Figure \ref{fig:rada}, show that our proposed model's multi-reference images performance is comparable to commercial solutions in several metrics, even with some advantages in dress consistency and human motion smoothness.

\subsection{Qualitative Results}

\begin{figure*}
  \centering
   \includegraphics[width=1\linewidth]{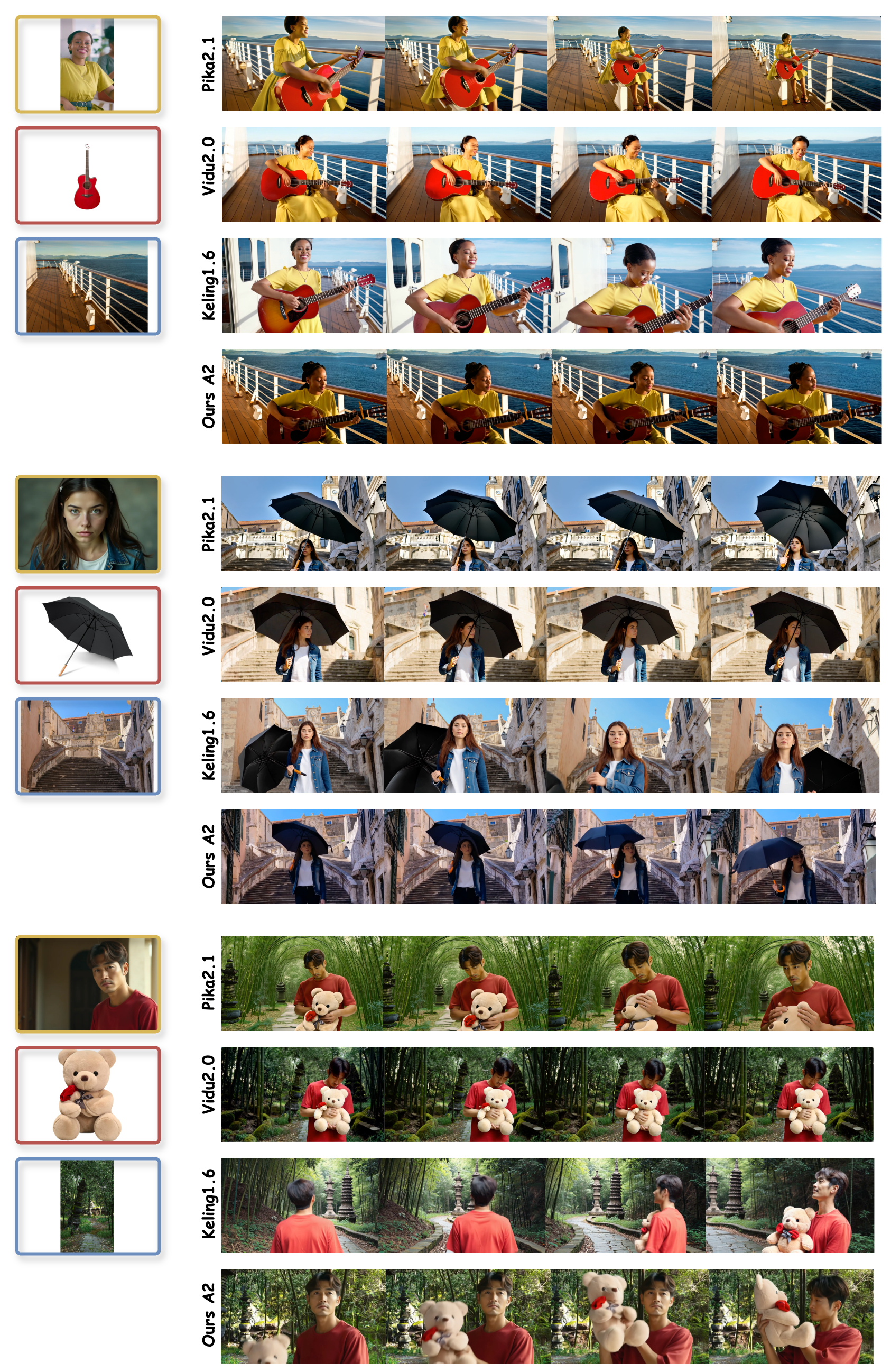}
   \caption{\textbf{Comparative results of elements-to-video generation with closed-source models.} We can see that our \texttt{SkyReels-A2} achieve achieves similar performance in composition and excels in the texture of light and shadow.}
   \label{fig:results2} 
\end{figure*}

\begin{figure*}
  \centering
   \includegraphics[width=1\linewidth]{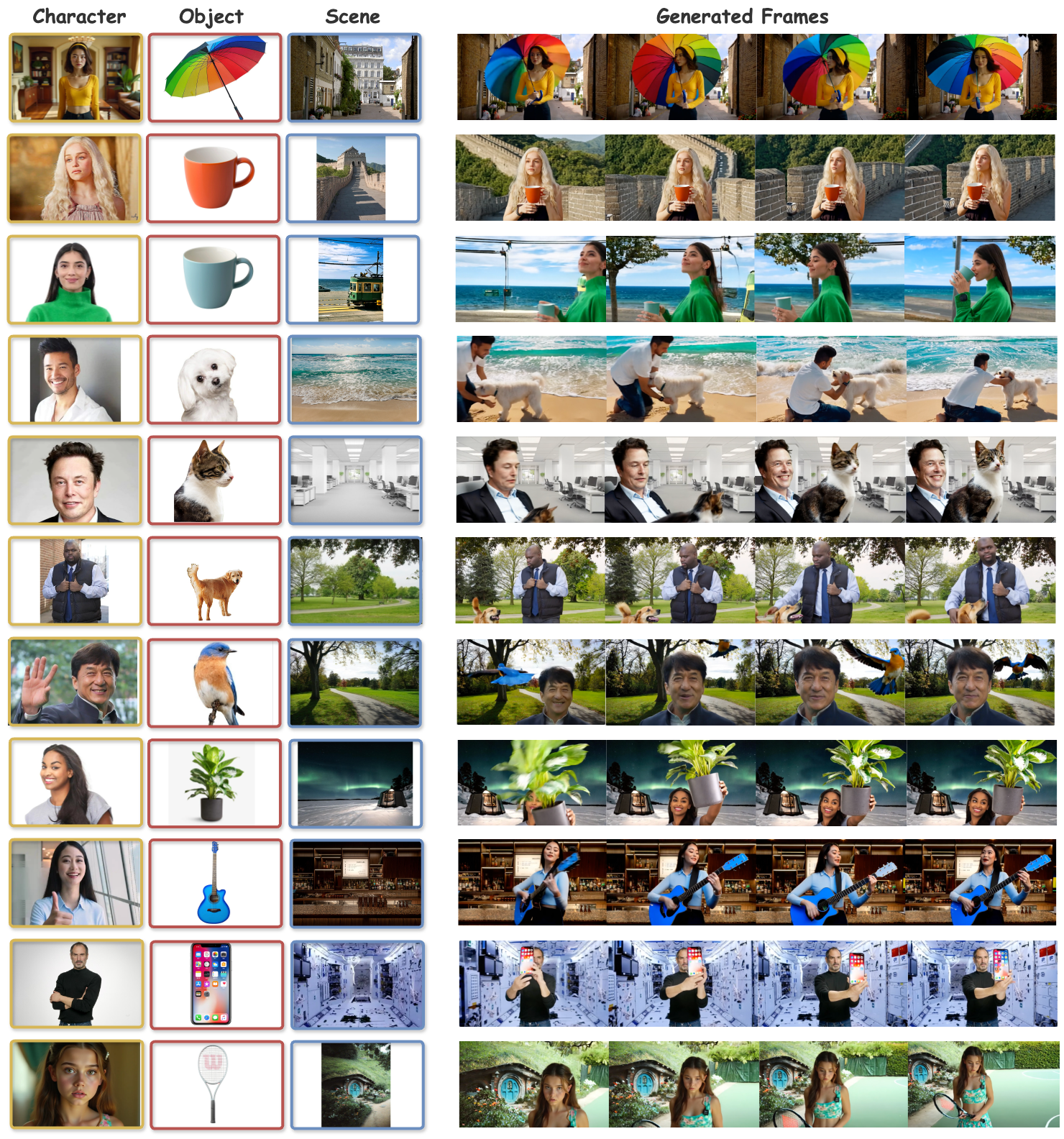}
   \caption{\textbf{More generated results of \texttt{SkyReels-A2}.} Our model has strong generalization ability and supports reference combinations of any subjects.}
   \label{fig:results_more} 
\end{figure*}

Here, we present the comparison results of several typical cases in Figure\ref{fig:results2}. Each generated video is displayed with four evenly sampled frames, including the first and last frames. Figures \ref{fig:results_more} show more results of generating multiple subject consistency from \texttt{SkyReels-A2}. 
It can be seen that Vidu and \texttt{SkyReels-A2} exhibit balanced performance in subject consistency, visual effect, and text response. Pika performs poorly in subject consistency with less motion. Keling has a notable has a noticeable mirror movement effect, from far to near or vice versa. We believe it is the impact caused by data distribution, similarly, our model was trained using more movie level data sources. Overall, our results are balanced across all dimensions, with particular advantage in subject character consistency and motion naturalness.

\subsection{Ablation Study}

In this section, we analyze the effect of different structure design of \texttt{SkyReels-A2} in detail. Here, we only change the ablation parts and control the residual model and training paradigm, i.e., training data and steps, seed, learning rate and so on, frozen. 

\begin{table}[t]
	\centering
    \caption{\textbf{Ablation study}. We compare different design space in \texttt{SkyReels-A2}, and select the best optimal algorithm, i.e., after VAE with frame repeat, cross-1, and 1:0 data mixing ration, in the final version. In between, $\alpha$ denote the single and multiple visual element data mixing ratio.}
	\begin{tabular}{lcccccc}
		\toprule
		\multirow{2}{*}{Methods} & \multicolumn{3}{c}{Composition Consistency} & \multicolumn{1}{c}{Visual Quality} & \multicolumn{1}{c}{Prompt Following} \\
		\cmidrule(lr){2-5} \cmidrule(lr){6-6}
		& ID $\uparrow$ & Obj. $\uparrow$ & Back. $\uparrow$ & Image Quality $\uparrow$ &  Similarity $\uparrow$ \\
		\midrule 
		Before VAE & 0.398 &0.809 &0.677 &0.683 &28.188 \\
        After VAE  & 0.388 &0.798 &0.697 &0.667 &28.186\\ 
        Frame No Repeat & 0.372 & 0.756 & 0.612 &0.642 & 26.232 \\
        \midrule
        Cross-2 & 0.362 & 0.732 &0.612 & 0.631 & 27.832\\
        Cross-1 & 0.398 &0.809 &0.677 &0.683 &28.188 \\
        Full &0.389 &0.798 &0.670 & 0.692 & 28.021 \\
        \midrule 
        $\alpha$ = 2:1 & 0.375 &0.772 &0.642 & 0.642  & 26.312 \\
        $\alpha$ = 1:1 & 0.386 &0.798 &0.665 & 0.671 & 27.561\\
        $\alpha$ = 0:1 & 0.398 &0.809 &0.677 &0.683 & 28.188 \\
		\bottomrule
	\end{tabular}
	\label{tab:abla}
\end{table}

\paragraph{Spacial feature combination.}

How to integrate spatial features is worth considering. Firstly, for the given multiple reference images, we attempted to (i) concatenate them in the original pixel space and then pass them through 3D VAE. Considering the compression of the temporal dimension, we also adopted the number of copies of the spatial dimension compression ratio in the pixel space (e.g., repeat for 4 times in Wan) to ensure that the image is not lost after being compressed by 3D VAE; (ii) Independently pass through 3D VAE and then concatenate the obtained features in latent space; (iii) We do not copy the compression temporal dimension, which cause more padding token in pixel space. 
The results can be seen in the first part of Table \ref{tab:abla}. We can see that no repeat makes a decreasing in image information, leads to a significant in composition consistency metrics. Referring to the generated examples, it can also be observed that there is an increase in the occurrence of subject loss without frame repeat.
Meantime, we also find that concatenate in latent space may cause interaction loss between different reference images, also leads to marginal performance degration.

\paragraph{Learnable parameters set.}

Generally speaking, the more parameters a model is trained on, the greater the damage to its original capabilities and the higher the data requirements. 
Based on the same training data collected, here, we consider different learnable parameter set strategies. Specifically, we consider: (i) only training the model parameters of cross-attention every two layers; (ii) training all the cross-attention layer learning mechanisms for all layers; (iii) Fine tuning of the entire video diffusion models. Note that we always include the image projection layers and patch embedding as learnable parts. According to the listed results in second part of Table \ref{tab:abla}, we find that only training every two cross-attention layers leads to a significant consistent performance loss, although reduce the memory request. Meantime, fine-tuning all the models helps to improve the image quality and makes the generated composition videos more naturalness, however, consider the performance and data scale balancing, training every cross-attention layers is optimal.

\paragraph{Training data mixing ratio.} 
Furthermore, we analyze the impact of data mixing on model performance. 
It is believed that integration of single ID data helps learn better representation in customization tasks \cite{huang2025conceptmaster}. 
Here, we first fixed the number of multi-subject data, and then increase the scale of single subject data. We denote single reference and multiple reference images ratio as $\alpha$ and modify the above ratios, setting them to 1:1, 2:1, and all multi subject data ratios. 
According to the results in the third part of Table \ref{tab:abla}, surprisingly, we found that adding single subject data did not improve the model performance in various composition scenarios. Therefore, we did not use mixing data in model training. We speculate that, similar to supervised fine-tuning, a moderate amount of high-quality text-reference-video data can better stimulate the controllability of the model without compromising the originally generated fluency.

\paragraph{Effect of inference acceleration and hyper-parameters.}

Considering that during training, the time step of schedule is set to 1000, and during inference, typically 30-50 is chosen. 
We need to sample inference step from 1000, and the in weighted schedule of controlled hyperparameter is flow shift. Here, we choose the best model and range this value from \{1, 3, 5, 8, 12\}. We find that the larger the value of flow shift selection, the more steps in early sampling, and the more reasonable the spatial structure, but the image details will gradually deteriorate. Taking into account the balance of motion and visual quality, we have chosen a value of 8 by default.

\subsection{Applications}


As an significant application of video generation models, we analyze their potential in music video creation and virtual E-commerce, with proposed \texttt{SkyReels-A2}. 
For music video creation, we can select instruments, such as guitars, and define scenes with more imagination (see Figure \ref{fig:results_more} third to last row) to generate seamless, creative sequences, if given music clips. 
Additionally, given product imagery (see Figure \ref{fig:results_more} second to last row), such as i-phone, we can design promotional content by placing celebrities in targeted scenarios. When paired with tailored voice-overs, this approach can effectively boost consumer purchasing intent. We will continue to pioneer innovative controllable generation scenarios to maximize the utility of video diffusion models.

\section{Related Works}

\subsection{Text-to-Video Generation}

The field of conditional video generation has seen remarkable progress, evolving from early approaches based on GANs and VAE to more advanced methodologies, such as diffusion-based models. Initial GAN-driven techniques \cite{vondrick2016generating, saito2017temporal, tulyakov2018mocogan, clark2019adversarial, yu2022generating} faced challenges in maintaining temporal consistency, often resulting in discontinuities between adjacent frames. To mitigate these issues, video diffusion models have been developed by adapting U-Net architectures—originally designed for text-to-image synthesis—thereby improving frame coherence.
Recent advancements have introduced diffusion Transformers \cite{peebles2023dit, fei2024scaling, fei2024flux, fei2023masked} and their extended variants \cite{U-ViT, fei2024scalable}, which replace conventional convolutional U-Net backbones \cite{ronneberger2015u} with fully Transformer-based architectures. These models commonly employ spatio-temporal attention mechanisms and comprehensive 3D attention \cite{peebles2023dit, lu2023vdt, ma2024latte, gao2024lumina}, significantly enhancing their capacity to model complex video dynamics and maintain frame continuity \cite{he2022latent, blattmann2023align, chen2023seine, girdhar2023emuvideo}. This shift towards Transformer-based architectures has contributed to improved scalability and facilitated more efficient parameter expansion.
In parallel, auto-regressive models \cite{yan2021videogpt, hong2022cogvideo, villegas2022phenaki, kondratyuk2023videopoet, xie2024show, liu2024mardini, fei2024diffusion, fei2019fast, fei2021partially} have leveraged discrete tokens to effectively capture temporal dependencies, demonstrating particular efficacy in generating extended video sequences \cite{yin2023nuwa, wang2023genlvideo, zhao2024moviedreamer, henschel2024streamingt2v, tan2024videoinfinity, zhou2024storydiffusion}.
This work investigates controlled application of video diffusion Transformers, with a specific focus on synthesizing compositions videos.

\subsection{Controllable Video Generation}

Large-scale, pre-trained text-to-video generation models \cite{esser2021taming, esser2024scaling} have been propelled by the integration of high-capacity textual representations \cite{radford2021learning, raffel2019exploring} and advanced attention-based architectures. These innovations have substantially contributed to achieving fine-grained control in the synthesis of temporally coherent video content \cite{li2025controlnet, mou2024t2i, bao2023latentwarp, qiu2024moviecharacter, peng2024controlnext, fei2024video, fei2023jepa}.
In identity-preserving tasks, tuning-based approaches typically involve adapting the pre-trained backbone to individual subjects at inference time \cite{Dreambooth, hu2021lora, textual_inversion, fei2023gradient, fei2022deecap}. While effective, these methods often incur considerable computational costs due to the need for per-identity fine-tuning. Although recent image and video generation models based on U-Net and DiT architectures have improved personalization capabilities \cite{Hyperdreambooth, multi-concept, dreamvideo, customvideo, motionbooth, Still-moving, ID-Animator, fang2024motioncharacter}, their reliance on continual model adaptation hinders scalability and limits broader deployment.
To mitigate the resource requirements inherent in tuning-based pipelines, the image generation domain has seen the emergence of more lightweight, tuning-free solutions \cite{Ip-adapter, instantid, UniPortrait, pulid, photomaker}. It is important to note that systems such as Keling\cite{Keling}, Vidu\cite{Vidu}, Pika\cite{Pika}, and MovieGen \cite{movie_gen} are closed-source. 
In contrast, we build upon the state-of-the-art video diffusion transformers and publicly release all technique details. Our approach facilitates the generation of coherent, high-fidelity, and editable videos while preserving subject-specific consistency, representing a significant advancement in scalable and controllable open video composition generation.

\section{Conclusion}

We introduce \texttt{SkyReels-A2}, a framework developed for elements-to-video generation that integrates multiple visual reference images. Generally, \texttt{SkyReels-A2} design the joint text-image model injection method from semantic and spacial perspectives based on existing video foundation models, to effectively learn cross-modal data form integration. The empirical results demonstrate that \texttt{SkyReels-A2} can deliver a composition synthesis that is not only of high quality and diversity but also offers robust editability and strong identity fidelity. We anticipate that our method will establish a new benchmark in the control video field, providing a replicable framework that can be adopted, extended, and optimized by future research efforts.

\section*{Acknowledges}

We would like to express our gratitude to Hao Zhang, Dixuan Lin, Zhiheng Xu for their help for constructing text-video dataset and infra, and Xin Sui, Binlu Zhang for model performance evaluation and insight feedback.



\bibliographystyle{ieee}
\bibliography{main}

\begin{table}[h]
    \small
    \centering
    \renewcommand{\arraystretch}{1.2} 
    \caption{Dimensions and Criteria for User Preference Studies.}
    \resizebox{\textwidth}{!}{
    \begin{tabular}{@{}ll@{}}
        \toprule
        \textbf{Evaluation Dimension} & \textbf{Evaluation Criteria and Details} \\ \midrule
        \multirow{3}{*}{\textbf{Face Consistency}} & \textbf{1:} Completely unable to identify as the same individual (facial features are unrelated to the reference image).  \\
        & \textbf{3:} Noticeable differences are present, but identification is still possible (e.g., variations in eye size or alterations in nose shape).  \\
        & \textbf{5:} All facial features are fully consistent (e.g., no differences in eye shape, nose curvature, or lip thickness). \\ \bottomrule
        \multirow{10}{*}{\textbf{Dress Consistency}} & \textbf{1:} The clothing is completely unrelated to the reference image (different style, color, and pattern). \\
           & For example, the reference wears traditional Hanfu, while the generated image shows swimwear.  \\
        & \textbf{2:} Significant style differences exist, with partial matches in color or pattern. \\
           & For instance, the reference wears a suit, and the generated image shows athletic wear, both in dark blue. \\
        & \textbf{3:} The styles are similar, but key elements have changed (e.g., long sleeves to short sleeves). \\
          & For example, the reference wears blue jeans, while the generated image features black casual pants with a similar cut. \\
        & \textbf{4:} The styles are consistent, with minor acceptable differences (e.g., color brightness or slight pattern scaling). \\
          & For example, the reference wears a white dress, and the generated image is in off-white, with matching hem lengths. \\
        & \textbf{5:} The clothing style, color, and pattern are identical (e.g., no differences in button placement or pattern details). \\
          & For instance, the reference wears a red plaid shirt, and the generated image matches the plaid pattern perfectly. \\ \bottomrule
        \multirow{10}{*}{\textbf{Background Consistency}} & \textbf{1:} The background is completely unrelated to the reference image (different scene, lighting, and objects). \\ & For example, the reference is in an office, while the generated image is a spacecraft. \\
        & \textbf{2:} There are significant scene type differences, with only partial matches in elements (e.g., color or material). \\ 
        & For instance, the reference is at the beach, while the generated image is a forest, both with green vegetation. \\
        & \textbf{3:} The scene types are similar, but key elements have changed (e.g., indoor to outdoor, day to night). \\
        & For example, the reference is in a kitchen, while the generated image shows a living room, with a consistent color scheme. \\
        & \textbf{4:} The scene types are consistent, with minor acceptable differences (e.g., lighting or object placement).\\ 
        & For instance, the reference is in a sunny park, while the generated image is cloudy, but vegetation and bench positions match.  \\
        & \textbf{5:} The scene elements are identical (e.g., no differences in furniture layout or decorations). \\
        & For example, the reference is in front of a library bookshelf, and the generated image matches the shelf arrangement exactly. \\ \bottomrule
        \multirow{3}{*}{\textbf{Object Consistency}} & \textbf{1:} Completely unrecognizable as the same object (features unrelated to the reference image). \\
        & \textbf{3:} Noticeable differences but still recognizable (e.g., changes in size or shape).\\
        & \textbf{5:} All object features are completely consistent (e.g., size, shape, and texture show no differences).\\ 
        \bottomrule
        \multirow{7}{*}{\textbf{Instruction Following}} & \textbf{1:} Fails to follow the instructions or only minimally aligns; significant deviation in theme and key elements.\\ 
        & \textbf{2:} Partially follows the instructions, but with notable deviations; key elements are present but lack completeness and accuracy. \\
        & \textbf{3:} Generally adheres to the instructions; main content and key elements are aligned, presenting the instruction adequately, \\ 
        & though with minor details missing or inaccurate.\\
        & \textbf{4:} Highly adheres to the instructions; accurately presents key content and elements without deviation. \\
        & \textbf{5:} Perfectly follows the instructions and extends beyond; the generated content adheres fully to the theme,\\
        & elements, and style, with enhancements that improve the overall effectiveness.\\ \bottomrule
        \multirow{10}{*}{\textbf{Rationality}} & \textbf{1:} Spatial relationships are entirely irrational; positions and interactions between subjects and backgrounds \\
        & or multiple subjects are contradictory, lacking any logical coherence.\\
        & \textbf{2:} Many spatial relationships among objects are flawed; interactions between subjects and backgrounds or among \\ 
        & multiple subjects show significant unnatural positioning and occlusion, severely impacting the video's realism and credibility. \\
        & \textbf{3:} Some spatial relationships are unreasonable; occasional issues with positioning, interactions, \\
        & or unnatural occlusion arise, but they do not significantly affect the overall presentation.\\
        & \textbf{4:} Only a few spatial relationships show slight deviations but are generally logical; interactions among \\ 
        & multiple subjects are reasonably coherent, and the overall output aligns closely with the instructions.\\
        & \textbf{5:} All positional, occlusion, and interaction relationships among objects are highly rational and natural, \\
        & with meticulous detail handling, resulting in an almost perfect overall presentation.\\ \bottomrule
        \multirow{5}{*}{\textbf{Subject Coherence}} & \textbf{1:} The main subject is completely distorted, or there are significant differences between frames. \\
        & \textbf{2:} Multiple areas or large sections show distortion.\\
        & \textbf{3:} Minor distortions in details, such as slight finger twisting when holding objects. \\
        & \textbf{4:} No distortion or issues with the character. \\
        & \textbf{5:} No distortion at all, with additional strengths like natural facial expressions, resulting in excellent overall quality.\\\bottomrule
         \multirow{10}{*}{\textbf{Motion Smoothness}} & \textbf{1:} Movement is very jerky; severe stuttering and noticeable jumps create a poor viewing experience with no coherent motion. \\
        & \textbf{2:} Clear stuttering and disjointed movement; transitions between actions are abrupt, \\
        & making most scenes feel unnatural and hindering content understanding.\\
        & \textbf{3:} Movement smoothness is average; occasional stutters or disconnections occur, but overall can still \\ & follow the motion rhythm, minimally impacting viewing of the main content.\\
        & \textbf{4:} Movement is relatively smooth and natural with no significant stuttering; \\
        & the motion feels fluid, and the trajectory is reasonable, providing a good overall experience.\\
        & \textbf{5:} Movement is extremely smooth and fluid, resembling natural motion with no stuttering or \\
        & disconnections, almost indistinguishable from real human movement.\\\bottomrule
        \multirow{9}{*}{\textbf{Aesthetic Quality}} & \textbf{1:} Extremely poor quality; severe distortion of the main subject or complete failure to meet prompt requirements,\\
        & rendering it unusable.\\
        & \textbf{2:} Overall quality is low; follows more than half of the instructions but presents incomplete content \\
        & (process without results or vice versa) and lacks basic aesthetics.\\
        & \textbf{3:} Moderate quality; conveys the prompt's information reasonably well with some detail flaws that \\
        & don’t impact the main storyline, exhibiting some aesthetic value but not immediately usable.\\
        & \textbf{4:} Good overall performance; fully meets instructions and presents the prompt information clearly without distortion \\ 
        & or missing elements, suitable for low-cost production.\\
        & \textbf{5:} Perfect overall performance.\\\bottomrule
        \multirow{7}{*}{\textbf{Video Quality}} & \textbf{1:} Severe quality issues; the image is blurry with noticeable pixelation or other major visual flaws, making it nearly unwatchable. \\
        & \textbf{2:} Poor image quality; blurriness is evident, and while the content can be vaguely recognized, \\ 
        & it significantly affects the viewing experience. \\
        & \textbf{3:} Average quality; minor flaws like slight blurriness or noise are present but do not hinder understanding of the main content. \\
        & \textbf{4:} Acceptable quality for viewing; clear images with no significant flaws, providing good visual effects on various devices. \\
        & \textbf{5:} Exceptional quality; professional-level image clarity without flaws in resolution, color, \\
        & contrast, or detail, suitable for high-quality display and dissemination.\\\bottomrule
    \end{tabular}
    }
\end{table}
\label{tab:dimensions}
\end{document}